\documentclass{article}

\usepackage{microtype}
\usepackage{graphicx}
\usepackage{subcaption}
\usepackage{booktabs} %

\usepackage{hyperref}

  \usepackage[preprint]{icml2026}

\usepackage{amsmath}
\usepackage{amssymb}
\usepackage{mathtools}
\usepackage{amsthm}

\usepackage[capitalize,noabbrev]{cleveref}

\theoremstyle{plain}

\theoremstyle{definition}

\theoremstyle{remark}

\usepackage[textsize=tiny]{todonotes}

\icmltitlerunning{KAN using Haar-like bases}

\usepackage[T1]{fontenc}
\usepackage[utf8]{inputenc}

\usepackage{cancel}

\usepackage{url}

\usepackage{tikz}
\usetikzlibrary{trees,positioning}

\usepackage{multirow}

\usepackage{tablefootnote}

\newcommand\R{\mathbb{R}}

\usepackage[english,russian]{babel}

\newcommand\SH{\textsf{\cancel{H}} }
\newcommand\KANI{KA\SH}

\newcommand\SPsi{\cancel\Psi}

\newcommand\KANSH{KAN/H}

\begin{document}
\selectlanguage{english}

\twocolumn[
  \icmltitle{\KANSH: Kolmogorov-Arnold Network Using Haar-like Bases}

  \icmlsetsymbol{equal}{*}

  \begin{icmlauthorlist}
    \icmlauthor{Susumu Katayama}{uom}
  \end{icmlauthorlist}

  \icmlaffiliation{uom}{Department of Engineering, University of Miyazaki, Miyazaki, Japan}

  \icmlcorrespondingauthor{Susumu Katayama}{skata@cs.miyazaki-u.ac.jp}

  \icmlkeywords{Kolmogorov-Arnold Network, wavelets, PATRICIA tree, function approximation}

  \vskip 0.3in
]

\printAffiliationsAndNotice{}  %

\begin{abstract}
Function approximation using Haar basis systems offers an efficient implementation when compressed via Patricia trees while retaining the flexibility of wavelets for both global and local fitting.
However, like B-spline-based approximations, achieving high accuracy in high dimensions remains challenging.
This paper proposes \KANSH, a variant of the Kolmogorov-Arnold Network (KAN) that uses a Haar-like hierarchical basis system with nonzero first-order derivatives, instead of B-splines.
We also propose a learning-rate scheduling method and a method for handling unbounded real-valued inputs, leveraging properties of linear approximation with Haar-like hierarchical bases.
By applying the resulting algorithm to function-approximation problems and MNIST, we confirm that our approach requires minimal problem-specific hyperparameter tuning.
 \end{abstract}

\section{Introduction}
The Kolmogorov-Arnold Network (KAN) \cite{liu2025kan} is a multilayer network constructed on the Kolmogorov-Arnold Theorem (KAT). KANs can approximate any continuous function with arbitrary accuracy and have potential applications in fields such as mathematics and physics. KANs are known for their ability to approximate nonlinear functions and are attracting attention in machine learning and data analysis.

KAT, the theoretical basis of KAN, states that ``any multivariate function can be approximated with arbitrary precision by sums of unary functions.'' Based on this theorem, KAN has a multi-layer network structure and can approximate complex multivariate functions by computing sums of unary functions at each layer.

KAN uses B-splines to approximate unary functions. B-splines are very useful basis functions for approximating continuous functions and achieve high accuracy, especially for unary functions. However, B-splines have problems such as high computational cost and learning difficulties:
\begin{itemize}
\item While the locality of each B-spline basis is useful for achieving high accuracy with minimal computational cost, it also reduces generalization performance during learning. In B-splines, there is a trade-off between accuracy and generalization, and one must find the optimal number of grid points for each problem.
\item Although B-splines are designed for functions on a predetermined finite domain, in multi-layer networks they cannot guarantee that hidden layer outputs remain within this domain. This necessitates ad-hoc approaches such as grid updates.
\end{itemize}

\KANSH (or \KANI in short), our proposed variant of KAN, solves the above problems by using a one-dimensional Haar-like hierarchical basis system.

One-dimensional Haar basis systems have the following desirable properties:
\begin{itemize}
\item it has a hierarchical structure containing both local and global bases, allowing highly accurate approximations while maintaining global generalization performance during learning;
\item an efficient algorithm \cite{katayama2000, dissertation} using an extended PATRICIA tree is known;
      for $n$ training samples and $p$-bit input precision,
	  the time required to update one sample is $O(\min \{\log n, p\})$, and the space complexity is $k\min\{n, 2^p\}$ bytes, where $k$ is a small integer. %
\item it can use the whole set of real numbers as the domain by directly using floating-point representations.
\end{itemize}

However, a network using only Haar bases cannot employ backpropagation because their derivatives are zero almost everywhere. Our proposed approach avoids this problem by using a Haar-like hierarchical basis system with simple piecewise-linear wavelets whose first derivatives are nonzero at specific points.
(Fig.~\ref{fig:wavelets}b)
We call this basis system the \emph{Slash-Haar} (or \SH{}) system.
The \SH{} system retains the good properties of the Haar system while enabling backpropagation with minimal loss of efficiency.
Moreover, by using Haar bases as global bases and \SH{} bases for local bases, the entire set of real numbers can be used as the domain.

This research also proposes learning-rate scheduling and a method for handling unbounded real-valued inputs, leveraging properties of linear approximation with \SH{} basis systems.
By applying the resulting algorithm to function approximation problems and MNIST, we confirm that minimal problem-specific hyperparameter tuning is required.
Specifically, as long as the learning rate is set to 1 or less, parameters did not diverge to infinity or NaN.

While this research focuses on obtaining a redundant and general KAN, the original KAN paper \cite{liu2025kan} simplifies KAN after obtaining a redundant KAN.
Since the same simplification method can be applied, we do not address it here.

The structure of this paper is as follows.
In Section~\ref{sec:preparation},
we review prior work on KAN and efficient implementations of function approximation using the Haar system.
Section~\ref{sec:proposed} presents the proposed method.
In Section~\ref{sec:slash_haar}, we propose using the \SH basis system and describe how to implement function approximation using \SH with a PATRICIA tree.
In Section~\ref{sec:realValueExtension}, we describe how to handle unbounded real-valued inputs.
In Section~\ref{sec:optimization}, we discuss the optimizer, particularly learning-rate scaling and scheduling methods tailored to the \SH basis system.
In Section~\ref{sec:kan_sh}, we explain how to construct and implement KAN with \SH.
In Section~\ref{sec:experiments}, we present experimental results on function-approximation problems and MNIST.
Finally, we conclude in Section~\ref{sec:conclusion}.

\section{Preparation}\label{sec:preparation}

\subsection{Kolmogorov-Arnold Network (KAN)}
The Kolmogorov-Arnold Network (KAN) is a multi-layer network for function approximation. Each layer of KAN computes its output as a sum of unary functions based on the Kolmogorov-Arnold Theorem. Specifically, the output $\mathbf{x}_{l+1}$ at layer $l+1$ is computed from the input $\mathbf{x}_l = (x_{l,1}, \ldots, x_{l,n_l})$ at layer $l$ as:
\begin{align}
 \mathbf{x}_{l+1} &= \Xi_l(\mathbf{x}_l) \label{eq:Xi} \\
 &= \left( \sum_{i=1}^{n_l} \xi_{l,1,i}(x_{l,i}), ..., \sum_{i=1}^{n_l}\xi_{l, n_{l+1}, i}(x_{l,i}) \right)
\end{align}
where $\xi_{l,j,i}$ is a unary function that takes $x_{l,i}$ (the output of the $i$-th unit at layer $l$) and outputs to the $j$-th unit of layer $l+1$.

By using $\Xi$ defined in Eq.~\ref{eq:Xi}, an $L$-layered KAN can be defined as follows:
\begin{equation}
 KAN = \Xi_{L-1} \circ \Xi_{L-2} \circ \cdots \circ \Xi_0
\end{equation}

In \cite{liu2025kan},
$\xi$ is expressed as a weighted sum of a unary B-spline and a basis function (typically SiLU) that serves as a residual connection.
These weights and the B-spline parameters are learned via backpropagation.
\subsubsection{KAN variants}
Since the release of the arXiv preprint in 2024 \cite{arXivKAN}, many variants of KAN have been proposed.
Many of them use basis functions other than B-splines.

Chebyshev KAN \cite{ChebKAN} uses Chebyshev polynomials instead of B-splines, achieving high accuracy with few parameters while controlling errors at interval boundaries.

Wav-KAN \cite{bozorgasl2024WavKAN} uses wavelets; the most effective derivative of Gaussian (DOG) wavelet resembles a smoothed \SH{} wavelet, indicating a connection to this work. However, Wav-KAN estimates the unary function of each edge nonlinearly by updating coefficients, shifts, and scales of a fixed number of wavelets. In contrast, the proposed method performs linear approximation over an enormous number of basis functions by leveraging the fact that most coefficients change in similar ways, achieving this with limited computational cost. Thus, the learning methods are fundamentally different.

\subsection{PATRICIA tree}
A PATRICIA tree \cite{PATRICIA, IntMap98} is a type of compressed binary search tree that reduces redundant nodes by storing index bit strings as integers. It has long been used to represent data structures indexed by integers (e.g., Haskell's IntMap) and for implementing TD($\lambda$) learning with the Haar system \cite{katayama2000, dissertation}. The Context Tree Weighting algorithm \cite{CTW} is also essentially a PATRICIA tree.

In a general binary tree, each node is either a leaf, a node with one child, or a node with two children. Nodes with only one child are redundant, as they merely record whether to go left or right. PATRICIA trees eliminate this redundancy by removing single-child nodes and directly connecting their parents to children, storing the sequence of left/right decisions as a bit string. This allows the tree to contain no single-child nodes. For $n$ leaf nodes, a PATRICIA tree has $2n-1$ nodes, using memory proportional to the number of samples---making it highly memory-efficient.

\subsection{Haar system}
The Haar system is an orthogonal basis set used to represent functions on the interval $[0, 1)$. It consists of wavelet functions and is widely used in signal processing, image processing, data compression, and noise removal.

While the Haar system can be extended to two or more dimensions, the number of bases grows exponentially with dimension, making high-precision approximation difficult in high dimensions. Therefore, this paper focuses on the one-dimensional Haar system and its variants.

Let $\Psi'$ be the Haar system. $\Psi'$ can be defined %
as follows:
\begin{align}
  \Psi'_{j,k}(x) &=
  \begin{cases}
    2^{\frac{j}2},  & \text{ if \ $2^{-j}k \leq x < 2^{-j}(k+\frac12)$;} \\
    -2^{\frac{j}2}, & \text{ if \ $2^{-j}(k+\frac12) \leq x < 2^{-j}(k+1)$;} \\
    0, & \text{ otherwise.}
  \end{cases}   \label{eq:typicalHaar}
\end{align}

When the domain is limited to the interval $[0,1)$, the Haar system has a hierarchical structure, which can be easily understood by thinking of it as a binary tree, as shown in Fig.~\ref{fig:Haar}.
In this case, $j$ in the index $(j,k)$ represents the depth, and $k$ represents the position at that depth. $j\ge0$ and $k\ge0$, and $k<2^j$ for $j>0$.
In particular, we can interpret $k$ as a binary number and use it as an index to traverse the tree. Alternatively, we can use $2^j + k$ as an index.
Here, we can take advantage of the following properties:
\begin{itemize}
\item the child nodes of $(j,k)$ (other than the root) have indices $(j+1, 2k)$ and $(j+1, 2k+1)$, obtained by appending one bit to $k$;
\item each basis at depth $j$ covers the domain $[0,1)$ without overlap;
\item the support of parent node $(j,k)$ is bisected into the supports of its child nodes;
\item all bases whose support contains $x$ lie on the path from the root to $x$ (when $x$ is interpreted as a binary index);
      the function value in Haar approximation is the weighted sum of these bases.
\end{itemize}

The Haar system defined by Eq.~\ref{eq:typicalHaar} is orthonormal, so function values have absolute values that grow exponentially with depth. However, for learning, this exponential growth is unnecessary; exponential decay is preferable as it enables generalization and convergence as an infinite series.
For this reason, \cite{katayama2000} and \cite{dissertation} use the following discounted bases instead, which apply the factor $\sqrt\beta$:
\begin{align}
  \MoveEqLeft \Psi_{j,k}(x) = \nonumber\\
  \MoveEqLeft
  \begin{cases}
    \sqrt{1-\beta}\beta^{\frac{j}2},  & \text{if \ $2^{-j}k \leq x < 2^{-j}(k+1)$;} \\
    -\sqrt{1-\beta}\beta^{\frac{j}2}, & \text{if \ $2^{-j}(k+1) \leq x < 2^{-j}(k+2)$;} \\
    0, & \text{ otherwise.}
  \end{cases}   \label{eq:Haar}
\end{align}

% Draw small plots of phi and psi on dyadic interval [k/2^j,(k+1)/2^j]
\newcommand{\plotHaarRoot}[2]{% #1=j, #2=k
  \begin{tikzpicture}[scale=1, baseline=-0.6ex]
	\plotHaarRootPrime{#1}{#2}
  \end{tikzpicture}
}
\newcommand{\plotHaarRootPrime}[2]{% #1=j, #2=k	
  \def\L{0.55}
  \pgfmathsetmacro{\left}{#2*\L/(2^#1)}
  \pgfmathsetmacro{\right}{(#2+1)*\L/(2^#1)}
  \pgfmathsetmacro{\mid}{(\left+\right)/2}
  % compute numeric fractions (reduced) for left/mid/right
  \pgfmathtruncatemacro{\leftnum}{#2}
  \pgfmathtruncatemacro{\leftden}{int(pow(2,#1))}
  \pgfmathtruncatemacro{\gleft}{gcd(\leftnum,\leftden)}
  \pgfmathtruncatemacro{\leftnumr}{int(\leftnum/\gleft)}
  \pgfmathtruncatemacro{\leftdenr}{int(\leftden/\gleft)}

  \pgfmathtruncatemacro{\midnum}{int(2*#2+1)}
  \pgfmathtruncatemacro{\middend}{int(pow(2,#1+1))}
  \pgfmathtruncatemacro{\gmid}{gcd(\midnum,\middend)}
  \pgfmathtruncatemacro{\midnumr}{int(\midnum/\gmid)}
  \pgfmathtruncatemacro{\middendr}{int(\middend/\gmid)}

  \pgfmathtruncatemacro{\rightnum}{int(#2+1)}
  \pgfmathtruncatemacro{\rightden}{int(pow(2,#1))}
  \pgfmathtruncatemacro{\gright}{gcd(\rightnum,\rightden)}
  \pgfmathtruncatemacro{\rightnumr}{int(\rightnum/\gright)}
  \pgfmathtruncatemacro{\rightdenr}{int(\rightden/\gright)}
    % apply horizontal stretch (center-preserving) to reduce left/right margins
    \pgfmathsetmacro{\s}{1.2} % stretch factor (1.0 = no stretch)
    \pgfmathsetmacro{\center}{\L/2}
    \pgfmathsetmacro{\leftvis}{\center + \s*(\left-\center)}
    \pgfmathsetmacro{\midvis}{\center + \s*(\mid-\center)}
    \pgfmathsetmacro{\rightvis}{\center + \s*(\right-\center)}
    % force node bounding box to remain the original width so outer rectangle stays same
    \path[use as bounding box] (-0.8*\L,-0.5) rectangle (1.6*\L,0.6);
    % axes (x-axis at y=0) drawn using stretched coordinates
  \pgfmathsetmacro{\visZero}{\center + \s*(0-\center)}
  \pgfmathsetmacro{\visL}{\center + \s*(\L-\center)}
  \draw[->] (\visZero,0) -- (\visL+0.2*\L,0) node[right,draw=none,xshift=-6pt]{\tiny $x$};
%  % compute j+1 and display Psi_{j+1,k}
%  \pgfmathtruncatemacro{\jplusone}{int(#1+1)}
  \draw[->] (\visZero,-0.3) -- (\visZero,0.3) node[above,draw=none]{\tiny $\Psi_{#1,#2}(x)$};
  % (removed support shading) 
  % phi: constant level (normalized, smaller)
  \draw[semithick,sharp corners] (\leftvis,0) -- (\leftvis,0.133333) -- (\rightvis,0.133333) -- (\rightvis,0);
    % mark left, mid, right on x-axis with numeric labels in local coord [0,1]
  % print reduced fractions (display integer if denominator==1)
  \ifnum\leftdenr=1
    \node[below,draw=none] at (\leftvis,0) {\tiny $\leftnumr$};
  \else
    \node[below,draw=none] at (\leftvis,0) {\tiny $\tfrac{\leftnumr}{\leftdenr}$};
  \fi
%   \ifnum\middendr=1
%     \node[below,draw=none] at (\mid,0) {\tiny $\midnumr$};
%   \else
%     \node[below,draw=none] at (\mid,0)  {\tiny $\tfrac{\midnumr}{\middendr}$};
%   \fi
  \ifnum\rightdenr=1
    \node[below,draw=none] at (\rightvis,0) {\tiny $\rightnumr$};
  \else
    \node[below,draw=none] at (\rightvis,0) {\tiny $\tfrac{\rightnumr}{\rightdenr}$};
  \fi
}

\newcommand{\plotHaarBasis}[2]{% #1=j, #2=k
  \begin{tikzpicture}[scale=1, baseline=-0.6ex]
	\plotHaarBasisPrime{#1}{#2}
  \end{tikzpicture}
}

\newcommand{\plotHaarBasisPrime}[2]{% #1=j, #2=k
	\def\L{0.55}
    \pgfmathsetmacro{\left}{#2*\L/(2^#1)}
      \pgfmathsetmacro{\right}{(#2+1)*\L/(2^#1)}
      \pgfmathsetmacro{\mid}{(\left+\right)/2}
      % compute numeric fractions (reduced) for left/mid/right
      \pgfmathtruncatemacro{\leftnum}{#2}
      \pgfmathtruncatemacro{\leftden}{int(pow(2,#1))}
      \pgfmathtruncatemacro{\gleft}{gcd(\leftnum,\leftden)}
      \pgfmathtruncatemacro{\leftnumr}{int(\leftnum/\gleft)}
      \pgfmathtruncatemacro{\leftdenr}{int(\leftden/\gleft)}

      \pgfmathtruncatemacro{\midnum}{int(2*#2+1)}
      \pgfmathtruncatemacro{\middend}{int(pow(2,#1+1))}
      \pgfmathtruncatemacro{\gmid}{gcd(\midnum,\middend)}
      \pgfmathtruncatemacro{\midnumr}{int(\midnum/\gmid)}
      \pgfmathtruncatemacro{\middendr}{int(\middend/\gmid)}

      \pgfmathtruncatemacro{\rightnum}{int(#2+1)}
      \pgfmathtruncatemacro{\rightden}{int(pow(2,#1))}
      \pgfmathtruncatemacro{\gright}{gcd(\rightnum,\rightden)}
      \pgfmathtruncatemacro{\rightnumr}{int(\rightnum/\gright)}
      \pgfmathtruncatemacro{\rightdenr}{int(\rightden/\gright)}
  \pgfmathsetmacro{\off}{0.2*\L}
  \pgfmathsetmacro{\yoff}{0.3*\L}
  % apply same horizontal stretch as phi for axes and function
  \pgfmathsetmacro{\s}{1.2}
  \pgfmathsetmacro{\center}{\L/2}
  \pgfmathsetmacro{\leftvis}{\center + \s*(\left-\center)}
  \pgfmathsetmacro{\midvis}{\center + \s*(\mid-\center)}
  \pgfmathsetmacro{\rightvis}{\center + \s*(\right-\center)}
  % preserve node bounding box so outer rectangle doesn't change
  \path[use as bounding box] (-0.8*\L,-0.5) rectangle (1.6*\L,0.6);
  % axes (x-axis at y=0) drawn using stretched coordinates
  \pgfmathsetmacro{\visZero}{\center + \s*(0-\center)}
  \pgfmathsetmacro{\visL}{\center + \s*(\L-\center)}
  \draw[->] (\visZero,0) -- (\visL+0.2*\L,0) node[right,draw=none,xshift=-6pt]{\tiny $x$};
  % compute j+1 and display Psi_{j+1,k}
  \pgfmathtruncatemacro{\jplusone}{int(#1+1)}
  \draw[->] (\visZero,-0.3) -- (\visZero,0.3) node[above,draw=none]{\tiny $\Psi_{\jplusone,#2}(x)$};
  % draw thick zero parts outside the support so [0,1) definition is visible
  \draw[semithick] (\visZero,0) -- (\leftvis,0);
  \draw[semithick] (\rightvis,0) -- (\visL,0);
  % draw step function: + on left, - on right (smaller amplitude) using stretched positions
  \draw[semithick,sharp corners] (\leftvis,0.133333) -- (\midvis,0.133333) -- (\midvis,-0.133333) -- (\rightvis,-0.133333);
  % add vertical connectors at the transition points for visual consistency
  \draw[semithick] (\leftvis,0) -- (\leftvis,0.133333);
  \draw[semithick] (\midvis,0.133333) -- (\midvis,0);
  \draw[semithick] (\midvis,0) -- (\midvis,-0.133333);
  \draw[semithick] (\rightvis,0) -- (\rightvis,-0.133333);
    % mark left, mid, right on x-axis with numeric labels
  % print reduced fractions (display integer if denominator==1)
  % place tick labels using stretched coordinates; keep outward offsets in visual coords
  \pgfmathsetmacro{\offvis}{\s*\off}
  \ifnum\leftdenr=1
    \node[below,draw=none] at (\leftvis-\offvis,0) {\tiny $\leftnumr$};
  \else
    \node[below,draw=none] at (\leftvis-\offvis,0) {\tiny $\tfrac{\leftnumr}{\leftdenr}$};
  \fi
  \ifnum\middendr=1
    \node[below,draw=none] at (\midvis,-\yoff) {\tiny $\midnumr$};
  \else
    \node[below,draw=none] at (\midvis,-\yoff)  {\tiny $\tfrac{\midnumr}{\middendr}$};
  \fi
  \pgfmathsetmacro{\offvisr}{\s*\off}
  \ifnum\rightdenr=1
    \node[below,draw=none] at (\rightvis+\offvisr,0) {\tiny $\rightnumr$};
  \else
    \node[below,draw=none] at (\rightvis+\offvisr,0) {\tiny $\tfrac{\rightnumr}{\rightdenr}$};
  \fi
}

% iffalse

% New robust selector: allow configurable on/off styles plus a
% backward-compatible default. Provide two user APIs:
%  - \drawHaarSelector{<bits>}            (uses defaults)
%  - \drawHaarSelectorStyles{<on>}{<off>}{<bits>}
% where <on> and <off> are TikZ key lists like "draw=black,thick" or
% "draw=black,very thin,dashed" and <bits> is an 8-char string.
% Default styles (can be redefined by the user before calling):
\def\HaarEdgeOnStyle{draw=black,thick}
\def\HaarEdgeOffStyle{draw=black,very thin,dashed}

% Simple wrapper: keep backward compatibility (single-braced bits)
\def\drawHaarSelector#1{%
  \drawHaarSelectorStyles{\HaarEdgeOnStyle}{\HaarEdgeOffStyle}{#1}%
}

% Main entry: user supplies on/off style strings and the 8-char bitmask
\def\drawHaarSelectorStyles#1#2#3#4#5{\drawHaarSelectorStylesSz#1#2#3#4\scriptsize#5}

\def\drawHaarSelectorStylesSz#1#2#3#4#5#6{%
  \edef\haartempSel{#6}% store bits
  \def\HOn{#3}% on-style token (TikZ key list)
  \def\HOff{#4}% off-style token
  \def\plotphi{#1}
  \def\plotpsi{#2}
  \def\fontsz{#5}
  \expandafter\drawHaarSelCharsWithStyles\haartempSel
}

% Internal: same logic as before but use \HOn / \HOff when building edge tokens
\def\drawHaarSelCharsWithStyles#1#2#3#4#5#6#7#8{%
  % compute subtree sums
  \pgfmathtruncatemacro{\SallS}{#1+#2+#3+#4+#5+#6+#7+#8}
  \pgfmathtruncatemacro{\SleftS}{#1+#2+#3+#4}
  \pgfmathtruncatemacro{\SrightS}{#5+#6+#7+#8}
  \pgfmathtruncatemacro{\SnTenLeftS}{#1+#2}
  \pgfmathtruncatemacro{\SnTenRightS}{#3+#4}
  \pgfmathtruncatemacro{\SnElevenLeftS}{#5+#6}
  \pgfmathtruncatemacro{\SnElevenRightS}{#7+#8}
  % define full edge tokens using the passed style tokens
  % define Edge tokens by fully expanding the style tokens (use \edef)
  \ifnum\SallS>0 \edef\EdgeRoot{edge from parent[\HOn]}\else \edef\EdgeRoot{edge from parent[\HOff]}\fi
  \ifnum\SleftS>0 \edef\EdgeNzerozeroLeft{edge from parent[\HOn]}\else \edef\EdgeNzerozeroLeft{edge from parent[\HOff]}\fi
  \ifnum\SrightS>0 \edef\EdgeNzerozeroRight{edge from parent[\HOn]}\else \edef\EdgeNzerozeroRight{edge from parent[\HOff]}\fi
  \ifnum\SnTenLeftS>0 \edef\EdgeNtenLeft{edge from parent[\HOn]}\else \edef\EdgeNtenLeft{edge from parent[\HOff]}\fi
  \ifnum\SnTenRightS>0 \edef\EdgeNtenRight{edge from parent[\HOn]}\else \edef\EdgeNtenRight{edge from parent[\HOff]}\fi
  \ifnum\SnElevenLeftS>0 \edef\EdgeNelevenLeft{edge from parent[\HOn]}\else \edef\EdgeNelevenLeft{edge from parent[\HOff]}\fi
  \ifnum\SnElevenRightS>0 \edef\EdgeNelevenRight{edge from parent[\HOn]}\else \edef\EdgeNelevenRight{edge from parent[\HOff]}\fi
  % leaf edges
  \ifnum#1>0 \edef\EdgeLzero{edge from parent[\HOn]}\else \edef\EdgeLzero{edge from parent[\HOff]}\fi
  \ifnum#2>0 \edef\EdgeLone{edge from parent[\HOn]}\else \edef\EdgeLone{edge from parent[\HOff]}\fi
  \ifnum#3>0 \edef\EdgeLtwo{edge from parent[\HOn]}\else \edef\EdgeLtwo{edge from parent[\HOff]}\fi
  \ifnum#4>0 \edef\EdgeLthree{edge from parent[\HOn]}\else \edef\EdgeLthree{edge from parent[\HOff]}\fi
  \ifnum#5>0 \edef\EdgeLfour{edge from parent[\HOn]}\else \edef\EdgeLfour{edge from parent[\HOff]}\fi
  \ifnum#6>0 \edef\EdgeLfive{edge from parent[\HOn]}\else \edef\EdgeLfive{edge from parent[\HOff]}\fi
  \ifnum#7>0 \edef\EdgeLsix{edge from parent[\HOn]}\else \edef\EdgeLsix{edge from parent[\HOff]}\fi
  \ifnum#8>0 \edef\EdgeLseven{edge from parent[\HOn]}\else \edef\EdgeLseven{edge from parent[\HOff]}\fi

  % draw the tree using these edge-token macros
  \begin{tikzpicture}[level distance=18mm,
    every node/.style={align=center, rectangle, draw=black,very thick,solid, rounded corners, inner sep=1.0pt, minimum width=19pt, font=\fontsz},
    edge from parent/.style={draw=black,very thin},
    level 1/.style={sibling distance=100mm},
    level 2/.style={sibling distance=70mm},
    level 3/.style={sibling distance=35mm},
    level 4/.style={sibling distance=17mm}
  ]
  \node {\plotphi{0}{0}}
    child { node {\plotpsi{0}{0}} \EdgeRoot
        child { node {\plotpsi{1}{0}} \EdgeNzerozeroLeft
          child { node {\plotpsi{2}{0}} \EdgeNtenLeft
            child { node {\plotpsi{3}{0}} \EdgeLzero }
            child { node {\plotpsi{3}{1}} \EdgeLone }
        }
          child { node {\plotpsi{2}{1}} \EdgeNtenRight
            child { node {\plotpsi{3}{2}} \EdgeLtwo }
            child { node {\plotpsi{3}{3}} \EdgeLthree }
        }
      }
        child { node {\plotpsi{1}{1}} \EdgeNzerozeroRight
          child { node {\plotpsi{2}{2}} \EdgeNelevenLeft
            child { node {\plotpsi{3}{4}} \EdgeLfour }
            child { node {\plotpsi{3}{5}} \EdgeLfive }
          }
          child { node {\plotpsi{2}{3}} \EdgeNelevenRight
            child { node {\plotpsi{3}{6}} \EdgeLsix }
            child { node {\plotpsi{3}{7}} \EdgeLseven }
        }
      }
    };
  \end{tikzpicture}
}

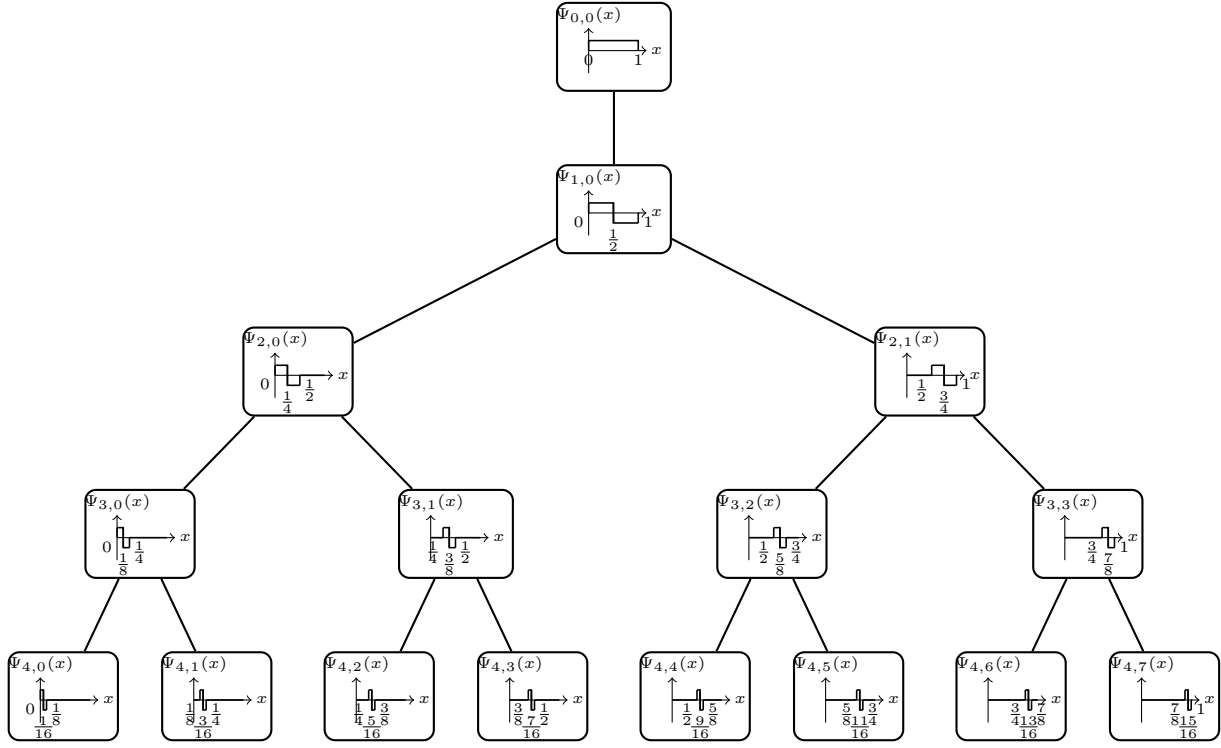
\begin{figure*}[tb]
  \centering

\begin{tikzpicture}[level distance=18mm, scale=1.2,
  every node/.style={align=center, rectangle, draw=black, thick, rounded corners, inner sep=1.0pt, minimum width=19pt, font=\scriptsize},
  edge from parent/.style={draw=black, thick},
  level 1/.style={sibling distance=100mm},
  level 2/.style={sibling distance=70mm},
  level 3/.style={sibling distance=35mm},
  level 4/.style={sibling distance=17mm}
]
\node (nm10) {\plotHaarRoot{0}{0} }
  child { node (n00) {\plotHaarBasis{0}{0} }
  child { node (n10) {\plotHaarBasis{1}{0}} 
  child { node (n20) {\plotHaarBasis{2}{0}} 
  child { node (n30) {\plotHaarBasis{3}{0}} }
  child { node (n31) {\plotHaarBasis{3}{1}} }
      }
    child { node (n21) {\plotHaarBasis{2}{1}
      } 
  child { node (n32) {\plotHaarBasis{3}{2}} }
  child { node (n33) {\plotHaarBasis{3}{3}} }
      }
    }
  child { node (n11) {\plotHaarBasis{1}{1}} 
      child { node (n22) {\plotHaarBasis{2}{2}} 
        child { node (n34) {\plotHaarBasis{3}{4}} }
        child { node (n35) {\plotHaarBasis{3}{5}} }
      }
      child { node (n23) {\plotHaarBasis{2}{3}} 
        child { node (n36) {\plotHaarBasis{3}{6}} }
        child { node (n37) {\plotHaarBasis{3}{7}} }
      }
    }
  };

\end{tikzpicture}

  \caption{Haar basis system on the unit interval. Each node (j+1,k) has support $[k/2^{j},(k+1)/2^{j}]$, children are obtained by halving the parent's interval.}
  \label{fig:Haar}
\end{figure*}

\subsection{Implementation of Haar system by an extended PATRICIA tree}\label{sec:haar-patricia}
Implementation of Haar bases using an extended PATRICIA tree is proposed by \cite{katayama2000} and \cite{dissertation}.\footnote{Although the term ``PATRICIA tree'' is not used in those papers, the trees defined there are extensions of PATRICIA trees.}
Note that the algorithms described in these papers are intended for use in reinforcement learning, and are therefore complex, requiring lazy computation of the eligibility traces.
In this paper, we perform simple function approximation and do not use eligibility traces.
Furthermore, these papers only deal with stochastic gradient ascent.
How to apply them to Adam will be described in 
 Section~\ref{sec:optimization}.
 
Each basis in the Haar system is arranged on a binary tree.
For each sample input, the basis set that includes that sample in its support can be collected by traversing down the binary tree using the binary representation of the input as an index.
One relatively naive implementation arranges the coefficients of each basis on a binary tree, and for each sample, computes the function value by computing the weighted sum of the coefficients on the path, and performs learning by updating the coefficients on the path.

A basis that has previously contained a sample in its support is called a visited basis,
and any other basis that has never previously contained a sample in its support is called an unvisited basis.
Since the coefficients of the unvisited bases should remain at their initial values,
it is sufficient to retain only the coefficients of the visited bases.
Therefore, the set of bases with values forms a tree similar to Figure \ref{fig:patricia-hyoro}a.

Furthermore, each basis on the consecutive edges without branches in Figure \ref{fig:patricia-hyoro}a should be updated in the same way. While the update in the positive or negative direction depends on whether the sample falls on the positive or negative part of the support, its absolute value is invariant (when multiplied by a depth-dependent weight).
Therefore, all consecutive sequences of edges can be compressed into a single edge that holds the coefficient and index information (Figure \ref{fig:patricia-hyoro}b). Note that while a normal PATRICIA tree only has values at the leaves, an extended PATRICIA tree for learning using a Haar basis set must also have values at the edges (or branches).

\def\skiptwo#1#2{~}
\def\writeRoot#1#2{$\Psi_{0,0}$}
\def\writePhi#1#2{\pgfmathtruncatemacro{\jplusone}{int(#1+1)}$\Psi_{\jplusone,#2}$}
\begin{figure*}[tb]
\centering
\begin{minipage}{0.45\textwidth}
    \centering
    \scalebox{0.5}{\drawHaarSelectorStylesSz\writeRoot\writePhi{draw=black,very thick}{draw=black,very thin,dashed}{\LARGE}{01100100}} \\
    (a)
\end{minipage}
\begin{minipage}{0.45\textwidth}
    \centering
    \begin{tikzpicture}
    \node {}
          child { node {}
                  edge from parent
                  child { node {}
                          edge from parent
                          child { node {} edge from parent }
                          child { node {} edge from parent }
                  }
                  child { node {} edge from parent }
          };
     \end{tikzpicture} \\
    (b)
\end{minipage}
\caption{Example of (a) visited Haar bases arranged on a binary tree when 3 different samples are input, and (b) its PATRICIA tree representation.}\label{fig:patricia-hyoro}
\end{figure*}
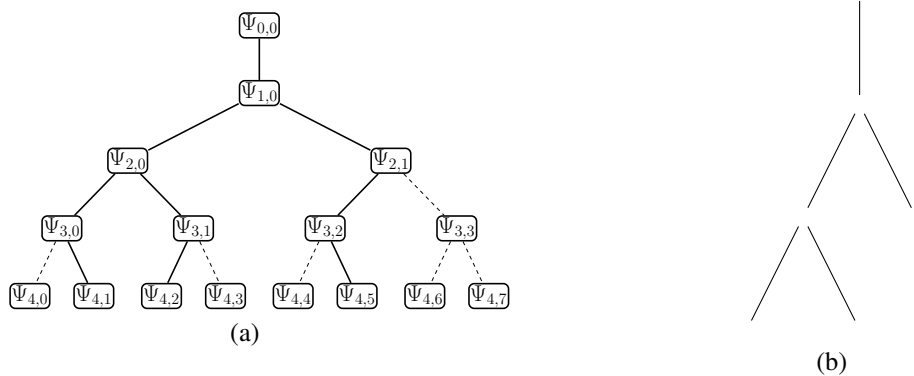

% Draw small plots of phi and psi on dyadic interval [k/2^j,(k+1)/2^j]
\newcommand{\plotSHRoot}[2]{% #1=j, #2=k
  \begin{tikzpicture}[scale=1, baseline=-0.6ex]
  \def\L{0.55}
  \pgfmathsetmacro{\left}{#2*\L/(2^#1)}
  \pgfmathsetmacro{\right}{(#2+1)*\L/(2^#1)}
  \pgfmathsetmacro{\mid}{(\left+\right)/2}
  % compute numeric fractions (reduced) for left/mid/right
  \pgfmathtruncatemacro{\leftnum}{#2}
  \pgfmathtruncatemacro{\leftden}{int(pow(2,#1))}
  \pgfmathtruncatemacro{\gleft}{gcd(\leftnum,\leftden)}
  \pgfmathtruncatemacro{\leftnumr}{int(\leftnum/\gleft)}
  \pgfmathtruncatemacro{\leftdenr}{int(\leftden/\gleft)}

  \pgfmathtruncatemacro{\midnum}{int(2*#2+1)}
  \pgfmathtruncatemacro{\middend}{int(pow(2,#1+1))}
  \pgfmathtruncatemacro{\gmid}{gcd(\midnum,\middend)}
  \pgfmathtruncatemacro{\midnumr}{int(\midnum/\gmid)}
  \pgfmathtruncatemacro{\middendr}{int(\middend/\gmid)}

  \pgfmathtruncatemacro{\rightnum}{int(#2+1)}
  \pgfmathtruncatemacro{\rightden}{int(pow(2,#1))}
  \pgfmathtruncatemacro{\gright}{gcd(\rightnum,\rightden)}
  \pgfmathtruncatemacro{\rightnumr}{int(\rightnum/\gright)}
  \pgfmathtruncatemacro{\rightdenr}{int(\rightden/\gright)}
    % apply horizontal stretch (center-preserving) to reduce left/right margins
    \pgfmathsetmacro{\s}{1.2} % stretch factor (1.0 = no stretch)
    \pgfmathsetmacro{\center}{\L/2}
    \pgfmathsetmacro{\leftvis}{\center + \s*(\left-\center)}
    \pgfmathsetmacro{\midvis}{\center + \s*(\mid-\center)}
    \pgfmathsetmacro{\rightvis}{\center + \s*(\right-\center)}
    % force node bounding box to remain the original width so outer rectangle stays same
    \path[use as bounding box] (-0.8*\L,-0.5) rectangle (1.6*\L,0.6);
    % axes (x-axis at y=0) drawn using stretched coordinates
  \pgfmathsetmacro{\visZero}{\center + \s*(0-\center)}
  \pgfmathsetmacro{\visL}{\center + \s*(\L-\center)}
  \draw[->] (\visZero,0) -- (\visL+0.2*\L,0) node[right,draw=none,xshift=-6pt]{\tiny $x$};
%  % compute j+1 and display Psi_{j+1,k}
%  \pgfmathtruncatemacro{\jplusone}{int(#1+1)}
  \draw[->] (\visZero,-0.3) -- (\visZero,0.3) node[above,draw=none]{\tiny $\SPsi_{#1,#2}(x)$};
  % (removed support shading) 
  % phi: constant level (normalized, smaller)
  \draw[semithick,sharp corners] (\leftvis,0.133333) -- (\rightvis,0);
    % mark left, mid, right on x-axis with numeric labels in local coord [0,1]
  % print reduced fractions (display integer if denominator==1)
  \ifnum\leftdenr=1
    \node[below,draw=none] at (\leftvis,0) {\tiny $\leftnumr$};
  \else
    \node[below,draw=none] at (\leftvis,0) {\tiny $\tfrac{\leftnumr}{\leftdenr}$};
  \fi
%   \ifnum\middendr=1
%     \node[below,draw=none] at (\mid,0) {\tiny $\midnumr$};
%   \else
%     \node[below,draw=none] at (\mid,0)  {\tiny $\tfrac{\midnumr}{\middendr}$};
%   \fi
  \ifnum\rightdenr=1
    \node[below,draw=none] at (\rightvis,0) {\tiny $\rightnumr$};
  \else
    \node[below,draw=none] at (\rightvis,0) {\tiny $\tfrac{\rightnumr}{\rightdenr}$};
  \fi
  \end{tikzpicture}
}

\newcommand{\plotSHBasis}[2]{% #1=j, #2=k
  \begin{tikzpicture}[scale=1, baseline=-0.6ex]
	\plotSHBasisPrime{#1}{#2}
  \end{tikzpicture}
}
\newcommand{\plotSHBasisPrime}[2]{% #1=j, #2=k
  \def\L{0.55}
    \pgfmathsetmacro{\left}{#2*\L/(2^#1)}
      \pgfmathsetmacro{\right}{(#2+1)*\L/(2^#1)}
      \pgfmathsetmacro{\mid}{(\left+\right)/2}
      % compute numeric fractions (reduced) for left/mid/right
      \pgfmathtruncatemacro{\leftnum}{#2}
      \pgfmathtruncatemacro{\leftden}{int(pow(2,#1))}
      \pgfmathtruncatemacro{\gleft}{gcd(\leftnum,\leftden)}
      \pgfmathtruncatemacro{\leftnumr}{int(\leftnum/\gleft)}
      \pgfmathtruncatemacro{\leftdenr}{int(\leftden/\gleft)}

      \pgfmathtruncatemacro{\midnum}{int(2*#2+1)}
      \pgfmathtruncatemacro{\middend}{int(pow(2,#1+1))}
      \pgfmathtruncatemacro{\gmid}{gcd(\midnum,\middend)}
      \pgfmathtruncatemacro{\midnumr}{int(\midnum/\gmid)}
      \pgfmathtruncatemacro{\middendr}{int(\middend/\gmid)}

      \pgfmathtruncatemacro{\rightnum}{int(#2+1)}
      \pgfmathtruncatemacro{\rightden}{int(pow(2,#1))}
      \pgfmathtruncatemacro{\gright}{gcd(\rightnum,\rightden)}
      \pgfmathtruncatemacro{\rightnumr}{int(\rightnum/\gright)}
      \pgfmathtruncatemacro{\rightdenr}{int(\rightden/\gright)}
  \pgfmathsetmacro{\off}{0.2*\L}
  \pgfmathsetmacro{\yoff}{0.3*\L}
  % apply same horizontal stretch as phi for axes and function
  \pgfmathsetmacro{\s}{1.2}
  \pgfmathsetmacro{\center}{\L/2}
  \pgfmathsetmacro{\leftvis}{\center + \s*(\left-\center)}
  \pgfmathsetmacro{\midvis}{\center + \s*(\mid-\center)}
  \pgfmathsetmacro{\rightvis}{\center + \s*(\right-\center)}
  % preserve node bounding box so outer rectangle doesn't change
  \path[use as bounding box] (-0.8*\L,-0.5) rectangle (1.6*\L,0.6);
  % axes (x-axis at y=0) drawn using stretched coordinates
  \pgfmathsetmacro{\visZero}{\center + \s*(0-\center)}
  \pgfmathsetmacro{\visL}{\center + \s*(\L-\center)}
  \draw[->] (\visZero,0) -- (\visL+0.2*\L,0) node[right,draw=none,xshift=-6pt]{\tiny $x$};
  % compute j+1 and display Psi_{j+1,k}
  \pgfmathtruncatemacro{\jplusone}{int(#1+1)}
  \draw[->] (\visZero,-0.3) -- (\visZero,0.3) node[above,draw=none]{\tiny $\SPsi_{\jplusone,#2}(x)$};
  % draw thick zero parts outside the support so [0,1) definition is visible
  \draw[semithick] (\visZero,0) -- (\leftvis,0);
  \draw[semithick] (\rightvis,0) -- (\visL,0);
  % draw step function: + on left, - on right (smaller amplitude) using stretched positions
  \draw[semithick,sharp corners] (\leftvis,0.133333) -- (\rightvis,-0.133333);
  % add vertical connectors at the transition points for visual consistency
  \draw[semithick] (\leftvis,0) -- (\leftvis,0.133333);
%  \draw[semithick] (\midvis,0.133333) -- (\midvis,0);
%  \draw[semithick] (\midvis,0) -- (\midvis,-0.133333);
  \draw[semithick] (\rightvis,0) -- (\rightvis,-0.133333);
    % mark left, mid, right on x-axis with numeric labels
  % print reduced fractions (display integer if denominator==1)
  % place tick labels using stretched coordinates; keep outward offsets in visual coords
  \pgfmathsetmacro{\offvis}{\s*\off}
  \ifnum\leftdenr=1
    \node[below,draw=none] at (\leftvis-\offvis,0) {\tiny $\leftnumr$};
  \else
    \node[below,draw=none] at (\leftvis-\offvis,0) {\tiny $\tfrac{\leftnumr}{\leftdenr}$};
  \fi
  \ifnum\middendr=1
    \node[below,draw=none] at (\midvis,-\yoff) {\tiny $\midnumr$};
  \else
    \node[below,draw=none] at (\midvis,-\yoff)  {\tiny $\tfrac{\midnumr}{\middendr}$};
  \fi
  \pgfmathsetmacro{\offvisr}{\s*\off}
  \ifnum\rightdenr=1
    \node[below,draw=none] at (\rightvis+\offvisr,0) {\tiny $\rightnumr$};
  \else
    \node[below,draw=none] at (\rightvis+\offvisr,0) {\tiny $\tfrac{\rightnumr}{\rightdenr}$};
  \fi
}

\section{Proposed method}\label{sec:proposed}
\subsection{Slash-Haar system}\label{sec:slash_haar}
We call the basis system $\SPsi$ the \emph{Slash-Haar basis system}, or simply the \SH system:
\begin{align}
  \MoveEqLeft \SPsi_{j,k}(x) = \nonumber\\
  \MoveEqLeft
 \begin{cases}
  \sqrt{1-\beta}\beta^{\frac{j}2}(1 + k - 2^j x),  & \text{if \ $2^{-j}k \leq x < 2^{-j}(k+2)$;} \\
  0,              & \text{otherwise.}
\end{cases}
\end{align}

\begin{figure}[tb]
  \centering
  \begin{minipage}{0.45\columnwidth}
	\centering
    \begin{tikzpicture}[scale=2, baseline=-0.6ex]
      \plotHaarBasisPrime{0}{0}
    \end{tikzpicture} \\
	{\small (a) Haar Wavelet}  %
  \end{minipage}
  \begin{minipage}{0.45\columnwidth}
	\centering
	\begin{tikzpicture}[scale=2, baseline=-0.6ex]
	  \plotSHBasisPrime{0}{0}
	\end{tikzpicture} \\
	{\small (b) Slash-Haar Wavelet}
  \end{minipage}
  \caption{Haar Wavelet and Slash-Haar Wavelet}
  \label{fig:wavelets}
\end{figure}
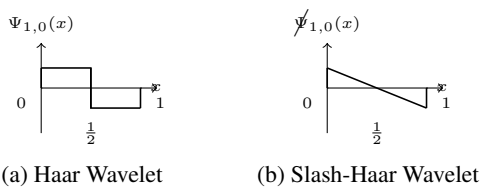

The \SH basis system $\SPsi$ is obtained by replacing the Haar Wavelet (Fig.~\ref{fig:wavelets}a) of the Haar system $\Psi$ with the \SH Wavelet (Fig.~\ref{fig:wavelets}b).

Each basis in the Haar system $\Psi$ can be expressed using $\SPsi$ as
\begin{align}
\Psi_{j,k}(x) = 2\SPsi_{j,k}(x) - \frac{\SPsi_{j+1,2k}(x)}{\sqrt\beta} - \frac{\SPsi_{j+1,2k+1}(x)}{\sqrt\beta}
\end{align}

Unlike the Haar system, each basis in the \SH system has a nonzero first derivative, which can be used for back-propagation:
\begin{align}
\frac{\partial}{\partial x}\SPsi_{j,k}(x) & = \begin{cases}
  - 2^j,  & \text{ if \ $2^{-j}k \leq x < 2^{-j}(k+2)$;} \\
  0,      & \text{ otherwise.}
\end{cases}
\end{align}
On the other hand, unlike higher-order B-spline approximations, \SH cannot be used to approximate higher-order derivatives due to the piecewise-linearity of $\Psi$.

The purpose of the variation from Haar to \SH is to enable approximation of the first derivative.
To achieve this, we need not use the whole $\SPsi$, but we can use $\Psi$ for shallow positions representing rough approximations and $\SPsi$ for deeper positions representing finer approximations.
In Section~\ref{sec:realValueExtension}, we use this idea of mixing $\Psi$ and $\SPsi$ in order to extend the domain from $(0,1)$ to $\R$.
We can also consider using the rectangular Haar scaling functions $\Phi$ instead of $\Psi$ for rough approximations near the root, in order to reduce the effect of discontinuities.

\subsubsection{Implementation of \SH system as a PATRICIA tree}

Since \SH bases have the same supports as the corresponding Haar bases, we can consider the PATRICIA tree for \SH with the same structure as in the case of Haar.
Note, however, that while the output $\Psi_{j,k}(x)$ of visited Haar nodes does not depend on the input $x$ except for the sign,
this is not true for \SH.
For this reason, the gradient $\partial y/\partial (\SPsi_{j,k}(x))$ of the output $y$ (a linear combination of visited \SH bases) also depends on the position of $x$ within the support of $\SPsi_{j,k}$.
As a result, if we apply the exact gradient method, coefficient values along consecutive only-child edges generally differ, which prevents compressing the binary coefficient tree into a PATRICIA tree.

Our approach does not stick to the exact gradient method, but instead uses a loosened hill-climbing method that permits inaccuracy in the absolute values provided that the sign for each dimension is correct.
More concretely, although it computes the output (and thus the error from the target) using \SH bases,
it updates the coefficients using the error as if it is using $\Psi$.
Our experimental results show that functions can be approximated correctly even in this way.

\newcommand\ffrac{{\mathrm{fraction}}}
\newcommand\fsgn{{\mathrm{sign}}}
\newcommand\fdbl{{\mathrm{dToUInt64}}}
\newcommand\fudbl{{\mathrm{udToUInt64}}}
\newcommand\gdbl{{\mathrm{dTo01}}}
\newcommand\gudbl{{\mathrm{udTo01}}}
\newcommand\fsig{{\mathrm{significand}}}
\newcommand\fexp{{\mathrm{exponent}}}
\newcommand\frnd{{\mathrm{round}}}
\subsection{Extension to unbounded real-valued inputs}\label{sec:realValueExtension}
$\Psi$ and $\SPsi$ have the domain of $[0,1)$.
When building a network like KAN, the domain should be extended to $(-\infty,\infty)$,
because inputs to layers other than the input layer can take any real value.
We consider the following candidate solutions:
\begin{enumerate}
 \item adaptively scale the domain (e.g. normalization)\cite{liu2025kan, bozorgasl2024WavKAN};
\label{item:scale}
 \item insert a (preferably monotonic) bounded mapping such as the logistic function \cite{shukla2024FAIR};
 \label{item:sig}
  \item define Haar and \SH over $(-\infty,\infty)$ and extend the tree lazily (on demand) whenever a sample falls outside the current domain; design an appropriate discount factor $\beta'$ for newly covered bases.
\label{item:lazy}
\item view the floating-point bit representation as the fixed-point representation in $[0,1)$ by the unsafe-cast operation; %
      roughly speaking, this corresponds to doing the following:
  \begin{enumerate}
   \item split the whole $(-\infty,\infty)$ region into $(-\infty,0)$ and $[0,\infty)$,
   \item partition each part into blocks by the values of the exponents, and
   \item use \SH within each block.
  \end{enumerate}
       This is a lightweight instance of Solution~\ref{item:sig}. \label{item:Floating}
\end{enumerate}
 This paper adopts Solution~\ref{item:Floating} (floating-point mapping), using Haar for the sign and exponent components.
 See Appendix~\ref{sec:ieee754} for implementation details. This approach is simple to implement and proved sufficiently accurate in our experiments.
  We also tried layer normalization (Solution~\ref{item:scale}) and the logistic function (Solution~\ref{item:sig}) in preliminary experiments, but they were not as good as Solution~\ref{item:Floating}.
Note also that using the logistic function can cause extreme accuracy degradation and loss of information (values saturate to $1$ or $0$)
when inputs slightly exceed $(-1,1)$.

\subsection{Optimization}\label{sec:optimization}

 When applying stochastic gradient methods, model-aware learning-rate scaling strongly affects stability and convergence speed. In hierarchical basis systems, commonly used rules such as the Linear Scaling Rule \cite{goyal2017accurate} and square-root scaling \cite{krizhevsky2014one} are unsafe because they assume similar update-frequency distributions across parameters. In our setting, deeper bases are visited far less frequently and follow different distributions. To address this, batch updates should be normalized by per-parameter visit counts (i.e., divide the sum of updates by the number of visits in the batch), which compensates for differing update frequencies.

 Learning-rate scheduling should depend on per-basis visit counts: frequently visited bases accumulate more information and therefore warrant smaller learning rates, while infrequently visited (deeper) bases keep larger learning rates for fine adjustments as training proceeds.

  Optimization by Adam \cite{kingma2017adam} can also be implemented with the same order of computational complexity as the stochastic gradient method.

Unlike SGA/SGD, Adam requires moment decay for unvisited nodes.
We can apply ``lazy updating'' \cite{katayama99, katayama2000},
where updates to the unvisited nodes are accumulated (like cache memory) at the roots of unvisited subtrees, until they are applied at once when they are visited.
This way, lazy updating effectively enables Adam implementation by only requiring computation for updating visited nodes and their sibling nodes.

Our preliminary experiments did not show any improvement by using Adam instead of SGD, possibly because we are using an approximate gradient, or because the samples were not batched.
In this paper, we only report results using SGD.

\subsection{\KANI: a variant of KAN using \SH instead of B-spline}\label{sec:kan_sh}
This paper proposes to replace the one-dimensional B-spline with the combination of Haar and \SH bases in the original KAN defined in \cite{liu2025kan}.
The good news is that our PATRICIA-based algorithm can handle multidimensional outputs with a single PATRICIA tree by storing a vector value (instead of a scalar) at each node, updating it at once, and outputting a vector at once.
Thus, for a KAN layer with $m$ inputs and $n$ outputs, only $m$ PATRICIA trees are required if each node stores an $\mathbb{R}^n$ vector; the number of tree traversals per forward pass is then $m$.

This $m$-way processing can be parallelized, although speedups are limited for small-scale KANs used in the function-approximation experiments of \cite{liu2025kan}.

Inputs to each layer are the raw floating-point numbers, as described in Section~\ref{sec:realValueExtension}.
For the input layer, however, it is possible to use fixed-point numbers when the bounds are known in advance, as in the MNIST handwritten digit dataset.

The proposed method works even without skip connections.
Also, simple skip connections such as Leaky ReLU can be obtained by just adjusting the initial coefficients when using fixed-point number inputs.
Our preliminary experiments suggest linear skip connections work better than no skip connections when using floating-point number inputs.

Unlike \cite{liu2025kan}, our proposed method does not require changing the basis set during training.
By using the floating-point expression directly, our method does not require changes in the basis set like grid update of \cite{liu2025kan}.
Moreover, by using a basis system that has both local and global bases, our method does not require changes in the basis set like the grid extension in \cite{liu2025kan}.

Although the memory consumption of our method increases as learning proceeds, it can be predictably bounded by the numeric precision. By choosing the precision based on available memory, our method can continue training indefinitely.

\subsection{Implementation issues for data-parallel environments}
Implementations of function approximation using B-spline can take advantage of GPU power.
On the other hand, while our method is advantageous when comparing performance on CPU because it is tree based,
its efficient implementation in data-parallel environments such as GPUs is not straightforward.

The binary tree can be easily implemented as a dense array on a GPU before compression to PATRICIA tree, unless the argument precision in bits is very high.
Since the accuracy experiments in \cite{liu2025kan} approach $10^{-7}$ precision, handling single-precision (\textasciitilde32-bit) accuracy is desirable.
However, implementation of \SH dealing with 32bit argument without compression to PATRICIA tree requires $2^{33}$ items in the array,
which means at least 32GB GPU memory.
Instead of a dense array, implementation of PATRICIA tree using sparse array is desired.

This paper mainly focuses on CPU implementations; exploiting GPU parallelism remains future work.

\section{Experiments}\label{sec:experiments}
We conducted experiments to evaluate the effectiveness of \SH and \KANI.
All experiments were performed in the environment described in Appendix~\ref{sec:expenv}.

\subsection{Experiment: accurate function approximation}\label{sec:func_approx}

To demonstrate the robustness of our method, we conducted all accuracy experiments from \cite{liu2025kan} (toy datasets, special functions, and Feynman datasets from Appendices N--P) using common hyperparameters.
Across all functions, we used the following common hyperparameters:

\begin{description}
\item[network structure:] $[n_{\mathrm{in}}, 5, 5, 1]$ ( $\langle \text{input} \rangle + \langle \text{hidden} \rangle \times 2 + \langle \text{output} \rangle$, where each hidden layer has 5 units.) for Section~\ref{sec:onlineExp}, \\
                           $[n_{\mathrm{in}}, 4, 4, 1]$ for
							Section~\ref{sec:offlineExp}.
\item[residual connection:] identity function
\item[optimizer:] SGD with fixed learning rate $\alpha=1$ for Section~\ref{sec:onlineExp}, SGD with cosine-scheduled learning rate starting with $\alpha=1$ and targeting the deadline of $10^7$ visits, based on the number of visits to each basis for
Section~\ref{sec:offlineExp}; 
\item[basis functions:] Haar with $\beta=1$ for the sign bit and the exponent 11 bits, \SH{} with $\beta=0.5$ for the significand 16 bits;
\item[dataset:] $10^7$ samples for training and $10^4$ samples for testing for each experiment for Section~\ref{sec:onlineExp}%
, $10^5$ samples for training and $10^4$ samples for testing for each experiment for Section~\ref{sec:offlineExp};
\item[random seeds:] we used a common seed for initializing the weights, a common seed for the training set, and a common seed for the test set throughout all experiments, which means that we did not cherry-pick random seeds for each experiment.
\item[input domain:] %
            $(0.1,0.9)$ for each input for Section~\ref{sec:onlineExp},
			$(0,1)$ for each input for Section~\ref{sec:offlineExp}.
            See details in the following discussion.
\end{description}

\subsubsection{Online learning experiment}\label{sec:onlineExp}
We first conducted experiments without learning-rate scheduling because the required number of epochs was unknown.

Although the input domain is arguably part of the task description rather than a hyperparameter, we could not find a corresponding description in \cite{liu2025kan}.
We first tried $(0,1)$ for each input dimension %
but since some Feynman functions diverge to infinity at some boundaries of the domain, it was difficult to learn them well enough to make the test-set RMSE converge to 0.
This is because, as long as learning progresses, there are always samples with extremely large values that dominate the RMSE calculation.
For this reason, we narrowed the input domain to $(0.1,0.9)$ in order to avoid these singular points in this experiment. %

Under the above conditions, the results are shown in the second column of Table~\ref{tbl:robustness} and Fig.~\ref{fig:robustness}.
Since parameters were common across experiments, all three categories could be run at one time.
Computation times were 11h 13m for the toy dataset, 10h 30m for the special functions, and 24h 23m for the Feynman dataset.
Note that these results were obtained without using a GPU.

Also note that II.11.7 and III.17.37 in the Feynman dataset are the same function. Since we use the same random seeds, the two curves overlap exactly.

From Table~\ref{tbl:robustness}, all functions except I.9.18 achieved RMSE $<10^{-2}$ and continue to improve (I.6.2b achieved this threshold only twice).
While I.9.18 learns slowly, likely due to problem difficulty, it still shows steady progress.

The original B-spline KAN lacks such hyperparameter robustness. We attempted to reproduce the original paper's results using its authors' GitHub code but encountered difficulties despite ad hoc tuning for each function.
Instabilities such as NaN have been reported not only for the original KAN %
but also for derived models like Chebyshev KAN \cite{shukla2024FAIR}. Thus, the stability of our method is a clear advantage.

\subsubsection{Offline learning experiment for dealing with singularities on the edges}\label{sec:offlineExp}
In the Feynman dataset, there are functions with singularities at the edges when the input domain is set to $(0,1)$, and these cannot be learned well with online learning with one epoch.
  In this section, we try to avoid this issue by
 \begin{itemize}
 \item limiting the training dataset size to $10^5$ and instead repeating training for $100$ epochs, and
 \item preventing the global bases that have already learned from being influenced by outliers by scheduling the learning rate for each basis.
 \end{itemize}

This experiment was conducted only for the Feynman dataset.
Also, since we did not use GPU acceleration, we used a smaller network structure to save time.

The third column of Table~\ref{tbl:robustness} and Fig.~\ref{fig:singularity} shows the results.
The proposed method never diverges, but for some functions with singularities at the edges, the final RMSE exceeds $1$, making it difficult to suppress the RMSE in cases where the function values have large absolute values.

\begin{table}[tb]
\centering
\caption{Results of function approximation %
 using the common hyperparameters, in RMSE.}\label{tbl:robustness}
\begin{tabular}{llll}
\hline
\multicolumn{3}{c}{Toy dataset} \\ \hline
Function                  & online             \\ \hline
Bessel J0                 & $1.02 \times 10^{-4}$ \\
2-ary fun\tablefootnote{$\exp (\sin \pi x + y^2)$}
                          & $4.68 \times 10^{-4}$ \\
$xy$                      & $7.84 \times 10^{-5}$ \\
100-ary fun\tablefootnote{$\exp (\sum_{i=1}^{100} \sin^2 ( \pi x_i / 2)  / 100)$}
                          & $8.82 \times 10^{-5}$ \\
4-ary fun\tablefootnote{$\exp( (\sin(\pi(x^2_1 + x^2_2)) + \sin(\pi(x^2_3+x_4^2))) / 2)$}
                          & $3.75 \times 10^{-3}$ \\
\hline
\multicolumn{3}{c}{Special functions} \\ \hline
Function                  & online \\ \hline
ellipjsn & $5.59 \times 10^{-5}$ \\
ellipkinc & $7.36 \times 10^{-5}$ \\
ellipeinc & $8.73 \times 10^{-5}$ \\
jv & $9.53 \times 10^{-5}$ \\
yv & $4.35 \times 10^{-4}$ \\
kv & $7.33 \times 10^{-4}$ \\
iv & $1.08 \times 10^{-4}$ \\
lpmv0 & $1.33 \times 10^{-4}$ \\
lpmv1 & $1.03 \times 10^{-4}$ \\
lpmv2 & $6.09 \times 10^{-5}$ \\
sph-harm01 & $1.46 \times 10^{-5}$ \\
sph-harm11 & $2.57 \times 10^{-5}$ \\
sph-harm02 & $3.77 \times 10^{-5}$ \\
sph-harm12 & $3.54 \times 10^{-5}$ \\
sph-harm22 & $6.88 \times 10^{-5}$ \\ \hline
\multicolumn{3}{c}{Feynman dataset} \\ \hline
Function  & online             & offline \\ \hline
I.6.2  & $1.02 \times 10^{-3}$ & $4.51 \times 10^{-2}$ \\
I.6.2b & $9.13 \times 10^{-3}$ & $5.05 \times 10^{-1}$ \\
I.9.18 & $1.05$                &  $4.61$ \\
I.12.11 & $8.75 \times 10^{-5}$ & $1.46 \times 10^{-4}$ \\
I.13.12 & $7.94 \times 10^{-4}$ & $3.69$ \\
I.15.3x & $4.29 \times 10^{-4}$ & $6.04 \times 10^{-1}$ \\
I.16.6 & $7.23 \times 10^{-5}$ & $1.15 \times 10^{-4}$ \\
I.18.4 & $1.12 \times 10^{-4}$  &  $5.63 \times 10^{-5}$ \\
I.26.2 & $1.55 \times 10^{-4}$  &  $1.29 \times 10^{-4}$ \\
I.27.6 & $7.08 \times 10^{-5}$  &  $6.89 \times 10^{-5}$ \\
I.29.16 & $2.20 \times 10^{-3}$ & $4.83 \times 10^{-4}$ \\
I.30.3 & $7.60 \times 10^{-5}$  & $5.61 \times 10^{-5}$ \\
I.30.5 & $1.70 \times 10^{-3}$  & $4.35 \times 10^{-3}$ \\
I.37.4 & $2.81 \times 10^{-4}$  & $3.32 \times 10^{-4}$ \\
I.40.1 & $8.07 \times 10^{-5}$  & $1.21 \times 10^{-4}$ \\
I.44.4 & $3.71 \times 10^{-4}$  & $1.58 \times 10^{-3}$ \\
I.50.26 & $1.52 \times 10^{-4}$ & $1.74 \times 10^{-4}$ \\
II.2.42 & $9.72 \times 10^{-5}$  & $2.24 \times 10^{-4}$ \\
II.6.15a & $4.26 \times 10^{-5}$ &$6.89 \times 10^{-5}$ \\
II.11.7 & $3.07 \times 10^{-4}$  & $4.81 \times 10^{-4}$ \\
II.11.27 & $1.36 \times 10^{-4}$ & $1.93 \times 10^{-4}$ \\
II.35.18 & $5.79 \times 10^{-5}$ &  $4.73 \times 10^{-5}$ \\
II.36.38 & $7.11 \times 10^{-4}$ & $8.82 \times 10^{-4}$ \\
II.38.3 & $2.00 \times 10^{-3}$  & $4.77$ \\
III.9.52 & $2.53 \times 10^{-4}$ & $3.18 \times 10^{-4}$ \\
III.10.19 & $6.09 \times 10^{-5}$ & $5.27 \times 10^{-5}$ \\
III.17.37 & $3.07 \times 10^{-4}$ &$4.81 \times 10^{-4}$ \\
\hline
\end{tabular}
\end{table}

\subsection{Experiment: MNIST}
To demonstrate that our algorithm applies to different tasks, we tested it on MNIST.

For the MNIST experiment, we used the following hyperparameters:
\begin{description}
\item[network structure:]  $[28\times28, 32, 1]$
\item[residual connection:] identity function
\item[optimizer:] SGD with cosine-scheduled learning rate starting with $\alpha=1$ and targeting the deadline of $30$ visits, based on the number of visits to each basis; 
\end{description}

\begin{figure}
 \centering
 \includegraphics[width=0.95\linewidth]{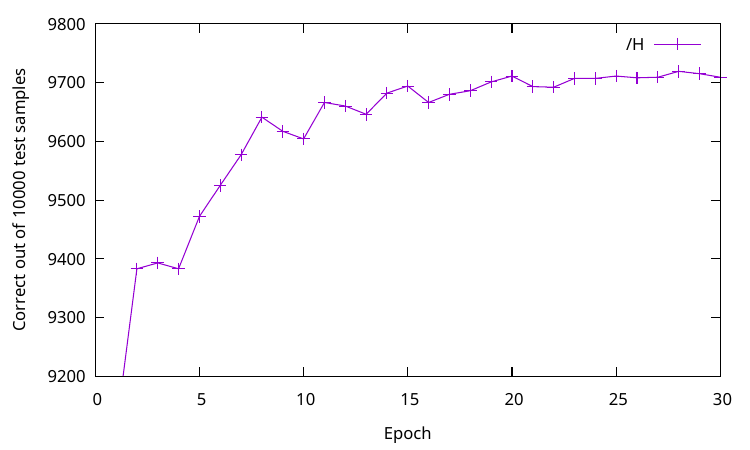}
 \caption{MNIST results.}\label{fig:MNIST}
\end{figure}

Fig.~\ref{fig:MNIST} shows the results of the MNIST experiment.

 We obtained similar results as those when using the Derivative of Gaussian (DOG) basis, which gave the best results in \cite{bozorgasl2024WavKAN}.
 The result graph is shown in Fig.~\ref{fig:MNIST}.

\section{Conclusions}\label{sec:conclusion}
We proposed \KANI, a KAN that uses Haar-like \SH bases and requires little problem-specific hyperparameter tuning.

\section{Future work}

As already noted, our current implementation assumes execution on CPU.
Accelerating the process by enabling it to run on a GPU is an urgent priority.
The current implementation does not efficiently deal with multi-core parallelism, and there is room for improving CPU implementation as well.

The proposed method is based on \cite{katayama99} and \cite{katayama2000}, which also work for TD($\lambda$)\cite{Sutton88},
and thus, it should be able to use eligibility traces even for \SH bases with similar computational complexity.
Furthermore, TD($\lambda$) was later integrated with the concepts of Q($\lambda$) \cite{Watkins89} and importance sampling \cite{importanceSampling} to lead to Retrace($\lambda$) \cite{retrace}.
Whether or not these algorithms can be implemented with similar computational complexity is also an interesting question.

\section*{Impact Statement}

This paper presents work whose goal is to advance the field of Machine
Learning. There are many potential societal consequences of our work, none
of which we feel must be specifically highlighted here.

\clearpage

\bibliographystyle{icml2026}
\bibliography{skatayama}

\newpage
\appendix

\section{Environment}
\label{sec:expenv}

We conducted experiments in the following environment:
\begin{description}
\item[OS]:       Ubuntu Jammy (22.04.2 LTS)
\item[CPU]:      Intel Core i9-12900, 4073.326MHz
\item[Compiler]: Glasgow Haskell Compiler version 9.6.7
\end{description}

\section{Results of the experiments in Section~\ref{sec:func_approx}}\label{sec:commonHyper}

\begin{figure*}[th]
\centering
\includegraphics[width=\textwidth]{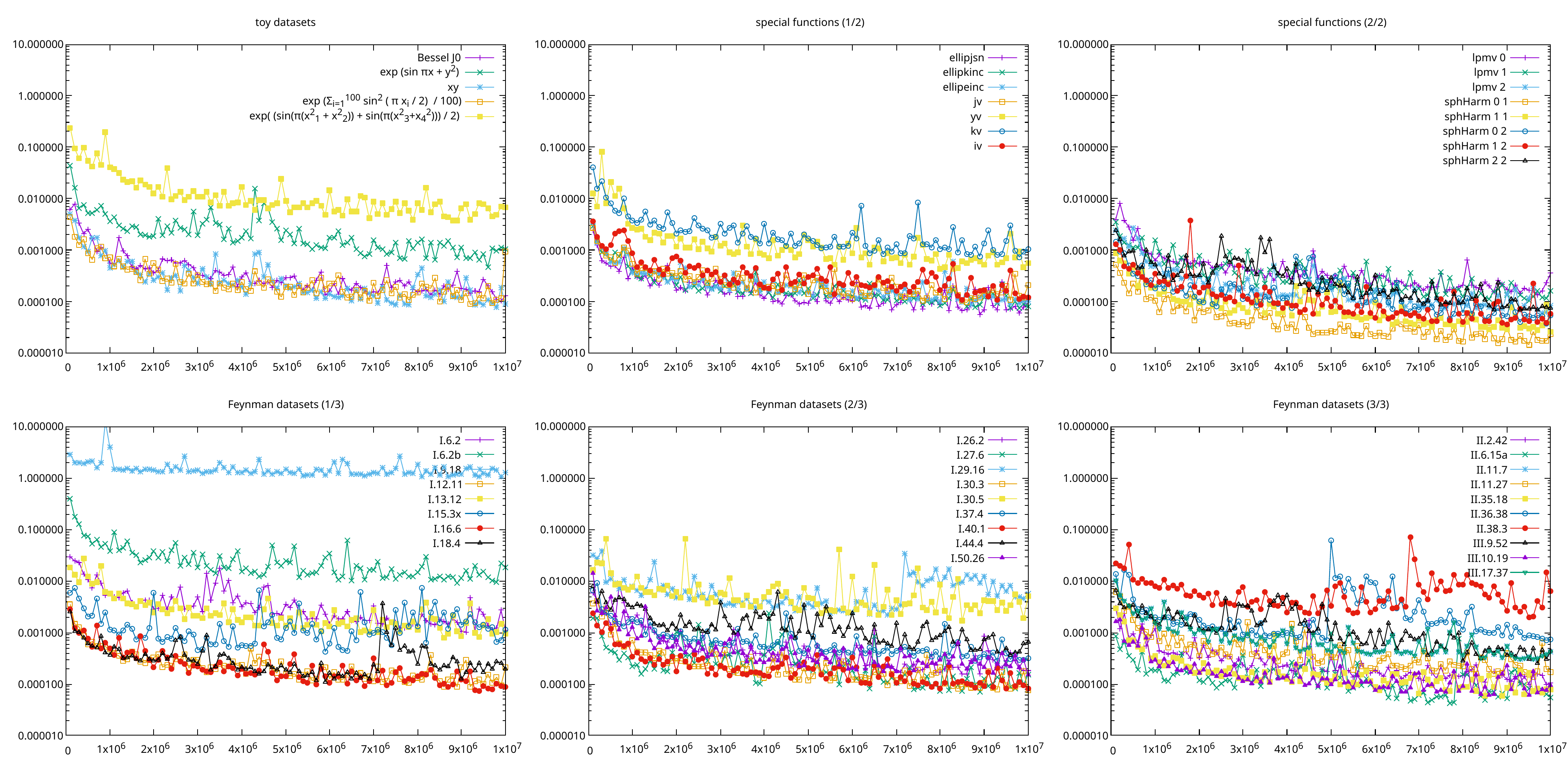}
\caption{Accuracy of online function approximation with input domain $(0.1,0.9)$ using the common hyperparameters, training steps vs. RMSE.}\label{fig:robustness}
\end{figure*}

\begin{figure*}[th]
\centering
\includegraphics[width=\textwidth]{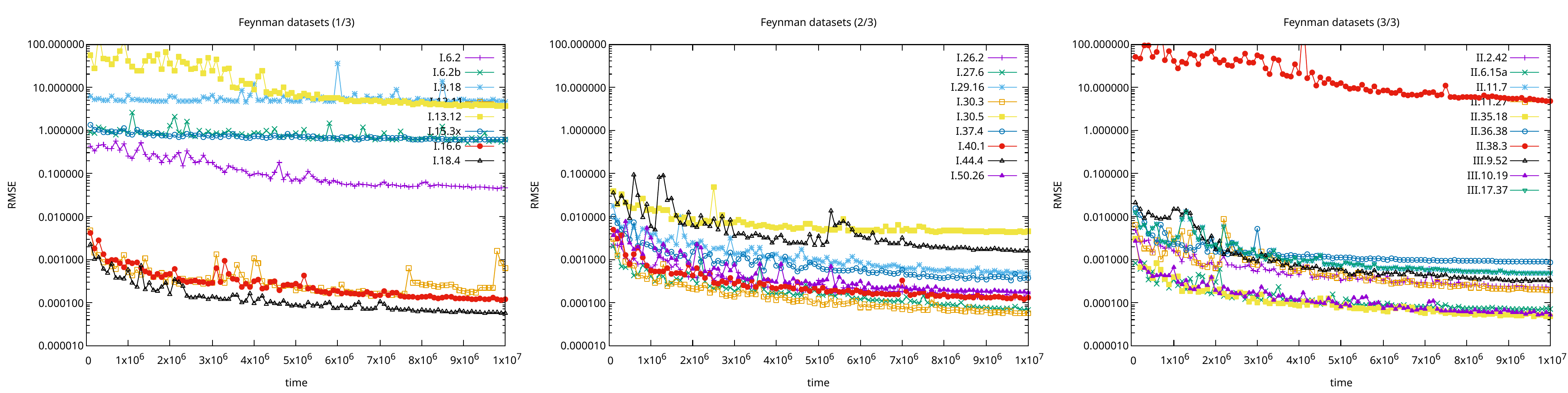}
\caption{Accuracy of offline function approximation with input domain $(0,1)$ using the common hyperparameters, training steps vs. RMSE.}\label{fig:singularity}
\end{figure*}

Figure~\ref{fig:robustness} shows results of online approximation of functions without singularities using the above hyperparameters for Section~\ref{sec:onlineExp}.
Figure~\ref{fig:singularity} shows results of offline approximation of functions with singularities on the edges using the above hyperparameters for Section~\ref{sec:offlineExp}.

\section{Floating-point conversion}
\label{sec:ieee754}

This section elaborates on the conversion from real numbers to $[0, 1)$ mentioned in Section~\ref{sec:realValueExtension}.

Currently, the floating-point representation widely adopted is the IEEE754 standard.
While it is not necessary to conform to this standard, for the purposes of explanation,
we will use IEEE754 double-precision real numbers.

A double-precision real number is defined as 1 bit for the sign, 11 bits for the exponent, 52 bits for the mantissa, and a bias of 1023.
Therefore, the function $\fdbl : \R \rightarrow \{0,...,2^{64}-1\}$ that converts a real number to a 64-bit unsigned integer can be expressed as follows:
\begin{align}
  \MoveEqLeft \fdbl(x) = \nonumber\\
  \MoveEqLeft
	 \left\{\begin{array}{l l}
                   \fudbl(x), & \text{ if $\fsig(x) \ge 0$;} \\
                   2^{63} + \fudbl(-x), & \text{ if $\fsig(x)<0$.}
     \end{array}\right. \\
  \MoveEqLeft \fudbl(x) = \nonumber\\
  \MoveEqLeft
 	 \left\{\begin{array}{r}
           \frnd(2^{52}\fsig(x)), \hfill \text{ if $\fexp(x) = -1023$;} \\
          2^{52}(\fexp(x)+1022) + \frnd(2^{53}\fsig(x)),
\\
		    \text{ ow.}
     \end{array}\right.
\end{align}
where $\fsig(x)$ is the significand of $x$, and $\fexp(x)$ the exponent of $x$.
They satisfy the following (in)equations:
\begin{align}
  \fsig(x)  &= 2^{-\fexp(x)}x \\
 |\fsig(x)| &< 1 \\
 |\fsig(x)| &\ge 1/2, \text{ if $\fexp(x) \ne -1023$}
\end{align}
Also, $\frnd$ rounds the fractional part of the argument in some way.
If you are using IEEE754-compliant arithmetic registers, this computation is not required for conversion; the unsafe cast operation is sufficient.

To convert a 64-bit unsigned integer value to a real value in $[0,1)$, simply divide it by $2^{64}$.
However, if it is rounded by $\frnd$, the derivative is $0$ and back-propagation is not possible.
For this reason, consider the continuous function $\gdbl : \R \rightarrow [0,1)$, assuming there is no $\frnd$:
\begin{align}
  \MoveEqLeft \gdbl(x) = \nonumber\\
  \MoveEqLeft
 \left\{\begin{array}{l l}
                   \gudbl(x), & \text{ if $\fsig(x) \ge 0$;} \\
                   2^{-1} + \gudbl(-x), & \text{ if $\fsig(x)<0$.}
 \end{array}\right. \label{eq:gdbl}\\
  \MoveEqLeft \gudbl(x) = \nonumber\\
  \MoveEqLeft
  \left\{\begin{array}{l}
           2^{-12}\fsig(x), \hfill \text{ if $\fexp(x) = -1023$;} \\
           2^{-12}(\fexp(x)+1022 + 2\fsig(x)), \hfill \text{ o.w.}
  \end{array}\right.
\end{align}

Except for discontinuous parts, we can assume the following equations:
\begin{align}
\frac{\mathrm{d}}{\mathrm{d} x}\fexp(x) &= 0  \\
\frac{\mathrm{d}}{\mathrm{d} x}\fsig(x) &= 2^{-\fexp(x)}
\end{align}
Thus,
\begin{align}
\frac{\mathrm{d}}{\mathrm{d}x}\gudbl(x) &= \left\{\begin{array}{l}
           2^{1011}, \hfill \text{ if $\fexp(x) = -1023$;} \\
           2^{-\fexp(x)-11}, \hfill \text{ otherwise.}
          \end{array}\right.
\end{align}
However, in practice, there will be no inconvenience in learning even if we do not specially treat the case of $\fexp(x) = -1023$. By combining this with the differentiation of Eq.~\ref{eq:gdbl}, we obtain
\begin{align}
\frac{\mathrm{d}}{\mathrm{d}x}\gdbl(x)
           &= \fsgn(\fsig(x)) 2^{-\fexp(x)-11} \label{eq:gdblprime}
\end{align}
for $\fsig(x)\ne0$.
Strictly speaking, Eq.~\ref{eq:gdblprime} becomes discontinuous when $\fsig(x)=0$, but we believe this does not cause problems in learning.

When generating samples, we discard any samples that contain NaN or $\pm\infty$. If $\pm\infty$ appears during the computation, during training, the sample is skipped as if it did not exist, and during testing, the computation is continued as is (using the unsafe cast value). The implementation is set up to display an error if NaN appears during the computation, but no NaN error has actually occurred.

\end{document}